\documentclass[conference, letterpaper]{IEEEtran}

\makeatletter
\def\endthebibliography{%
	\def\@noitemerr{\@latex@warning{Empty `thebibliography' environment}}%
	\endlist
}
\makeatother

\usepackage{cite}

\usepackage[pdftex]{graphicx}
\graphicspath{ {figures/} }
\DeclareGraphicsExtensions{.pdf,.jpeg,.png}

\usepackage{amsmath}
\usepackage{amssymb}

\usepackage{tabularx}
\usepackage{multirow}
\usepackage{booktabs}

\usepackage[caption=false,font=footnotesize]{subfig}

\usepackage{url}

\newcommand{\norm}[1]{\left\lVert #1 \right\rVert}

\begin{document}
\title{Visual Global Localization with\\a Hybrid WNN-CNN Approach}

\author{
	\IEEEauthorblockN{
		Avelino Forechi$^{a,b}$\IEEEauthorrefmark{1},
		Thiago Oliveira-Santos$^{b}$\IEEEauthorrefmark{3},
		Claudine Badue$^{b}$\IEEEauthorrefmark{2},
		Alberto Ferreira De Souza$^{b}$\IEEEauthorrefmark{2}
	}
	\and
	\and
	\IEEEauthorblockA{
		\hspace{1cm}$^{a}$Coordenadoria de Engenharia Mec\^anica\\
		\hspace{1cm}Instituto Federal do Esp\'irito Santo\\
		\hspace{1cm}Aracruz, Esp\'irito Santo, Brazil, 29192--733\\
		\hspace{1cm}\IEEEauthorrefmark{1}avelino.forechi@ifes.edu.br
	}	
	\and
	\IEEEauthorblockA{
		$^{b}$Departamento de Inform\'atica -- Centro Tecnol\'ogico\\
		Universidade Federal do Esp\'irito Santo\\
		Vit\'oria, Esp\'irito Santo, Brazil, 29075--910\\
		\IEEEauthorrefmark{2}\{alberto, claudine\}@lcad.inf.ufes.br, \IEEEauthorrefmark{3}todsantos@inf.ufes.br
	}
}

\maketitle

\begin{abstract}
Currently, self-driving cars rely greatly on the Global Positioning System (GPS) infrastructure, albeit there is an increasing demand for alternative methods for GPS-denied environments. One of them is known as place recognition, which associates images of places with their corresponding positions. We previously proposed systems based on Weightless Neural Networks (WNN) to address this problem as a classification task. This encompasses solely one part of the global localization, which is not precise enough for driverless cars. Instead of just recognizing past places and outputting their poses, it is desired that a global localization system estimates the pose of current place images. In this paper, we propose to tackle this problem as follows. Firstly, given a live image, the place recognition system returns the most similar image and its pose. Then, given live and recollected images, a visual localization system outputs the relative camera pose represented by those images. To estimate the relative camera pose between the recollected and the current images, a Convolutional Neural Network (CNN) is trained with the two images as input and a relative pose vector as output. Together, these systems solve the global localization problem using the topological and metric information to approximate the current vehicle pose. The full approach is compared to a Real-Time Kinematic GPS system and a Simultaneous Localization and Mapping (SLAM) system. Experimental results show that the proposed approach correctly localizes a vehicle 90\% of the time with a mean error of 1.20m compared to 1.12m of the SLAM system and 0.37m of the GPS, 89\% of the time.
\end{abstract}


% \fi
\IEEEpeerreviewmaketitle

\section{Introduction}

\begin{figure}[ht]
	\includegraphics[height = 3.5cm, width = 0.5\textwidth]{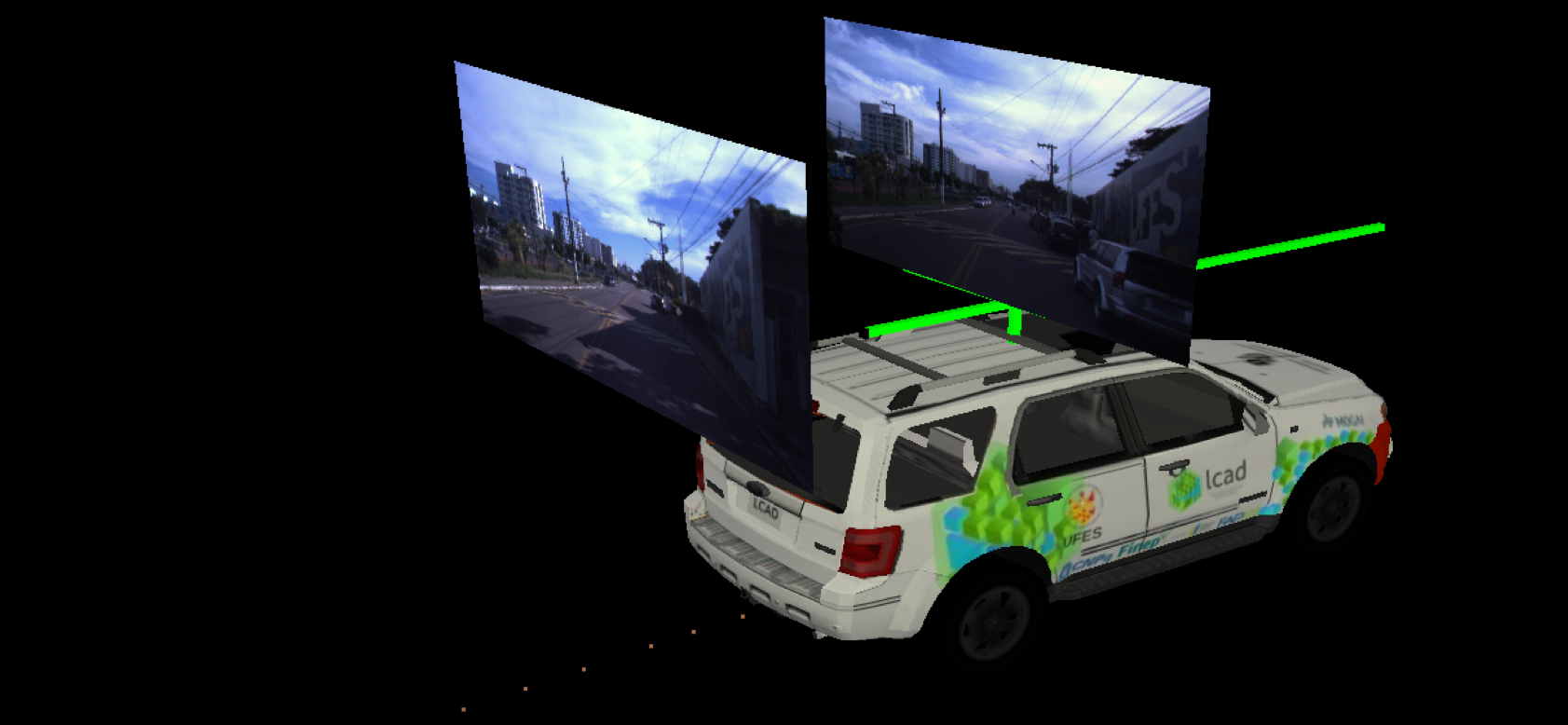}
	\caption[]{Illustration of IARA 3D viewer depicting two images (the left one is a recollected image and the right one is the current image) in their actual world positions and a orange dot trail indicating IARA's previous positions.} 
	\label{fig-IARA} 
\end{figure}

The Global Positioning System (GPS) has been widely used in ground vehicle positioning. When used in conjunction with Real-Time Kinematic (RTK) data or other sensors, such as Inertial Measurement Units (IMU), it can achieve a military-grade precision even in the positioning of urban vehicles. Although its widespread use in urban vehicle navigation, it suffers from signal unavailability in cluttered environments, such as urban canyons, under a canopy of trees and indoor areas. Such GPS-denied environments require alternative approaches, like solving the global positioning problem in the space of appearance, which involves computing the position given images of an in-vehicle camera. This problem is commonly approached as a visual place recognition problem and it is often modeled as a classification task \cite{Forechi2016, LyrioJunior2014}. However, that solves just the first part of the global localization problem, given that it learns past place images and returns the most similar past location and not its current position. A complementary approach to solve the global localization problem would be to compute a transformation from the past place to the current one given their corresponding images, as illustrated in Figure~\ref{fig-IARA}. This problem differs in some aspects from Visual Odometry (VO) \cite{Nister2004} and Structure from Motion (SfM) \cite{Huang1994} problems. Although all of them can be characterized as visual localization methods, estimating a relative camera pose is more general than the other two. VO computes motion from subsequent image frames, while the relative camera position method computes motion from non-contiguous image frames. SfM computes motion and structure of rigid objects based on feature matching at different times or from different camera viewpoints, while the relative camera position method does not benefit from the structure of rigid objects, most of the time, given that the camera motion is roughly orthogonal to the image plane. In addition to the place recognition task previously addressed with Weightless Neural Networks (WNN), the relative camera pose estimation is approached here using a Convolutional Neural Network (CNN) in order to regress the relative pose between past and current image views of a place. The full WNN-CNN approach is compared to a Real-Time Kinematic GPS system and a Visual Simultaneous Localization and Mapping (SLAM) system \cite{LyrioJunior2015}. Experimental results show that the proposed combined approach is able to correctly localize an autonomous vehicle 90\% of the time with a mean error of 1.20m compared to 1.12m of a Visual SLAM system and to 0.37m of the GPS, 89\% of the time. 

%This paper is organized as follows. After this introduction, in Section~\ref{sec:rel_work} it is presented an overview of related works, and in Section~\ref{sec:system} the proposed approach to solve the global localization problem using a hybrid WNN-CNN approach. In Section~\ref{sec:methodology}, it is described the experimental setup and methodology. Experimental results are presented and discussed in Section~\ref{sec:results}, leading to the conclusions and directions for future work in Section~\ref{sec:conclusion}.

\section{Related Work} \label{sec:rel_work}

In the following, we discuss how the global localization task is addressed in the literature with regards to place recognition and visual localization systems. Place recognition systems serve a large number of applications such as robot localization \cite{Engelson1994, Mur-Artal2015, LyrioJunior2015}, navigation \cite{Matsumoto1996, Cummins2007, Milford2009}, loop closure detection \cite{Lynen2015, Labbe2013, Galvez-Lopez, Hou2015}, to geo-tagged services \cite{WeyandGoogle, Lin, Hays2008}. Visual localization refers to the problem of inferring the 6 Degree of Freedom (DoF) camera pose associated with where images were taken. 

\subsubsection{Visual Place Recognition}

Herein, we discuss just the latest approaches to the problem. For a comprehensive review please refer to \cite{Lowry2015}. 
In \cite{Chen2014a}, authors presented a place recognition technique based on CNN, by combining the features learned by CNN's with a spatial and sequential filter. They employed a pre-trained network called Overfeat \cite{Sermanet2013} to extract the output vectors from all layers (21 in total). For each output vector, they built a confusion matrix from all images of training and test datasets using Euclidean distance to compare the output feature vectors. Following they apply a spatial and sequential filter before comparing the precision-recall curve of all layers against FAB-MAP \cite{Cummins2008} and SeqSLAM \cite{Milford2012}. It was found in \cite{Chen2014a} that middle layers 9 and 10 perform way better (85.7\% recall at 100\% precision) than the 51\% recall rate of SeqSLAM. 
In \cite{LyrioJunior2014, Forechi2016}, our group employed two different WNN architecture to global localize a self-driving car. Our methods have linear complexity like FAB-MAP and do not require any a priori pre-processing (e.g. to build a visual vocabulary). It is achieved with one method, called VibGL, an average pose precision of about 3m in a few kilometers-long route. VibGL was later integrated into a Visual SLAM system named VIBML \cite{LyrioJunior2015}. To this end, VibGL stores landmarks along with GPS coordinates for each image during mapping. VIBML performs position tracking by using stored landmarks to search for corresponding ones in currently observed images. Additionally, VIBML employs an Extended Kalman Filter (EKF)\cite{Thrun2005} for predicting the robot state based on a car-like motion model and corrects it using landmark measurement model\cite{Thrun2005}. VIBML was able to localize an autonomous car with an average positioning error of 1.12m and with 75\% of the poses with an error below 1.5m in a 3.75km path \cite{LyrioJunior2015}.

\subsubsection{Visual Localization}

In \cite{Song2015}, authors proposed a system to localize an input image according to the position and orientation information of multiple similar images retrieved from a large reference dataset using nearest neighbor and SURF features \cite{Bay2008}. The retrieved images and their associated pose are used to estimate their relative pose with respect to the input image and find a set of candidate poses for the input image. Since each candidate pose actually encodes the relative rotation and direction of the input image with respect to a specific reference image, it is possible to fuse all candidate poses in a way that the 6-DoF location of the input image can be derived through least-square optimization. Experimental results showed that their approach performed comparably with civilian GPS devices in image localization. Perhaps applied to front-facing camera movements as is the case here, their approach might not work properly as most of images would lie along a line causing the least-square optimization to fail.
In \cite{Agrawal2015}, the task of predicting the transformation between pair of images was reduced to 2D and posed as a classification problem. For training, they followed Slow Feature Analysis (SFA) method \cite{Chopra2005} by imposing the constraint that temporally close frames should have similar feature representations disregarding either the camera motion and the motion of objects in the scene. This may explain the authors' decision to treat visual odometry as a classification problem since the adoption of SFA should discard "motion features" and retain the scale-invariant features, which are more relevant to the problem as classification problem than to the original regression problem. 
Kendall et. al. \cite{Kendall2015a} proposed a monocular 6-DoF relocalization system named PoseNet, which employs a CNN to regress a 6-DoF camera pose from a single RGB image in an end-to-end manner. The system obtains approximately 3m and $6\deg$ accuracy for large scale (500 x 100m) outdoor scenes and 0.6m and $10\deg$ accuracy indoors. Its most salient features relate greatly to the environment structure since camera moves around buildings and most of the time it faces them.
The task of locating the camera from single input images can be modeled as a Structure of Motion (SfM) problem in a known 3D scene. The state-of-the-art approaches do this in two steps: firstly, projects every pixel in the image to its 3D scene coordinate and subsequently, use these coordinates to estimate the final 6D camera pose via RANSAC. In \cite{Brachmann2017}, the authors proposed a differentiable RANSAC method called DSAC in an end-to-end learning approach applied to the problem of camera localization. The authors have achieved an increase in accuracy by directly minimizing the expected loss of the output camera pose estimated by the DSAC.
In \cite{Oliveira}, the authors proposed a metric-topological localization system, based on images, that consists of two CNN's trained for visual odometry and place recognition, respectively, which predictions are combined by a successive optimization. Those networks are trained using the output data of a high accuracy LiDAR-based localization system similar to ours. The VO network is a Siamese-like CNN, trained to regress the translational and rotational relative motion between two consecutive camera images using a loss function similar to \cite{Kendall2015a}, which requires an extra balancing parameter to normalize the different scale of rotation and translation values. For the topological system part, they discretized the trajectory into a finite set of locations and trained a deep network based on DenseNet \cite{Huang2017a} to learn the probability distribution over the discrete set of locations.

\subsubsection{Visual Global Localization}

Summing up, a place recognition system capable of correctly locating a robot through discrete locations serves as an anchor system because it limits the error of the odometry system that tends to drift. By recalling that our place recognition system stores pairs of image-pose about locations, we just need to estimate a 6-DoF relative camera pose, considering as input the recollected image-pose from the place recognition system and a live camera image. The 6-DoF relative pose applied to the recollected camera pose takes the live camera pose in the global frame of reference. The Place Recognition system presented here is based on a WNN such as the one first proposed in \cite{LyrioJunior2014}, whilst the Visual Localization system is a new approach based on a CNN architecture similar to \cite{Handa2016}. 
In the end, our hybrid WNN-CNN approach does not require any additional system to merge the results of the topological and metric systems as the output of one subsystem is input to the other. Our relative camera pose system was developed concurrently to \cite{Melekhov2017} and differs in network architecture, loss function, training regime, and application purpose. While they regress translation and quaternion vectors, we regress translation and rotation vectors; while they use the ground-truth pose as supervisory signal in $L^2$-norm with an extra balancing parameter \cite{Kendall2015a}; we employ the $L^2$-norm on 3D point clouds transformed by the relative pose; while they use transfer learning on a pre-trained Hybrid-CNN \cite{Zhou2014} topped with two fully-connected layers for regression, we train a Fully Convolutional Network \cite{Long2015} from scratch. Finally, their system application is more related to Structure from Motion, where images are object-centric and ours is route-centric. We validated our approach on real-world images collected using a self-driving car while theirs were validated solely using indoor images from a camera mounted on a high-precision robotic arm. 

\section{Visual Global Localization with a Hybrid WNN-CNN Approach} \label{sec:system}

This section presents the proposed approach to solve the global localization problem in the space of appearance in contrast to traditional Global Positioning System (GPS) based systems. The proposed approach is twofold: (i) a WNN to solve place recognition as a classification problem and (ii) a CNN to solve visual localization as a metric regression problem. Given a live camera image, the first system recollects the most similar image and its associated pose, whilst the latter compares the recollected image to the live camera image in order to output a 6D relative pose. The final outcome is an estimation of the current robot pose, which is composed by the relative pose, given by system (ii), applied to the recollected image pose, given by system (i).

\subsection{Place Recognition System}\label{sec:place_recog}

Our previous place recognition system \cite{LyrioJunior2014}, named VibGL, employed a WNN architecture designed to solve image-related classification problems. VibGL employs a committee of Virtual Generalized Random Access Memory (VG-RAM) WNN units, called VG-RAM neurons, for short, which are individually responsible for representing binary patterns extracted from the input and associating with their corresponding label. VG-RAM neurons are organized in layers and can also serve as input to further layers in the architecture. As a machine learning method, VG-RAM has a supervised training phase that stores pairs of inputs and labels, and a test phase that compares stored inputs to new ones. Its testing-time increases linearly with the number of training samples. To overcome this problem, one can employ the Fat-Fast VG-RAM neuron proposed in \cite{Forechi2015} for faster neuron memory search, which is leveraged by an indexed data structure with a sub-linear runtime that assumes uniformly distributed patterns. 

Our latest place recognition system \cite{Forechi2016}, named SABGL, demonstrated that taking as input a sequence of images is more accurate than taking a single-image as VibGL. The reason is that a sequence of images provides temporal consistency. Although SABGL demonstrated better classification performance than VibGL, in this paper, we will further experiment with VibGL, but in a different scenario, where multiple similar images of a place are acquired over time and used for training. In this context, we are interested in exploiting data spatial consistency. 

\subsection{Visual Localization System}\label{sec:visual_localize}

In this section, it is described the system proposed to solve the visual localization problem by training a Siamese-like CNN architecture to regress a 6-DoF pose vector.

\subsubsection{Architecture}

The CNN architecture adopted here is similar to the one proposed by Handa et al. \cite{Handa2016}. Their architecture takes inspiration from the VGG-16 network \cite{Simonyan2014} and uses $3\times3$ convolutions in all but the last two layers, where $2\times1$ and $2\times2$ convolutions are used to compensate for the $320\times240$ resolution used as input, as opposed to the $224\times224$ used in the original VGG-16. Also, the top fully connected layers were replaced by convolutional layers, which turned the network into a Fully Convolutional Network (FCN) \cite{Long2015}.

The network proposed by Handa et al. \cite{Handa2016} takes in a pair of consecutive frames, $I_t$ and $I_{t+1}$, captured at time instances $t$ and $t+1$, respectively, in order to learn a 6-DoF visual odometry pose. The CNN adopted here, takes in a pair of image frames distant up to $d$ meters from each other. More specifically, the siamese network branches are fed with pairs of keyframes $I_K$ and live frames $I_L$, in which the relative distance of a live frame to a keyframe is up to $d$ meters. The network's output is a 6-DoF pose vector, $\delta_{pred}$, that transforms one image coordinate frame to the other. The first three components of $\delta_{pred}$ correspond to the rotation and the last three to the translation.

Similarly to the architecture proposed by Handa et al. \cite{Handa2016}, the one adopted here fuses the two siamese network branches earlier, in order to ensure that spatial information is not lost by the depth of the network. Despite the vast majority of CNN architectures alternate convolution and max-pooling layers, it is possible to replace max-pooling layers for a larger-stride convolution kernel, without loss in accuracy on several image recognition benchmarks \cite{Springenberg2014}. Based on this finding and seeking to preserve spatial information through out the network, there is no pooling layers in the network adopted here.

\begin{figure}[ht]
	\includegraphics[height=6cm, width = 0.5\textwidth,keepaspectratio]{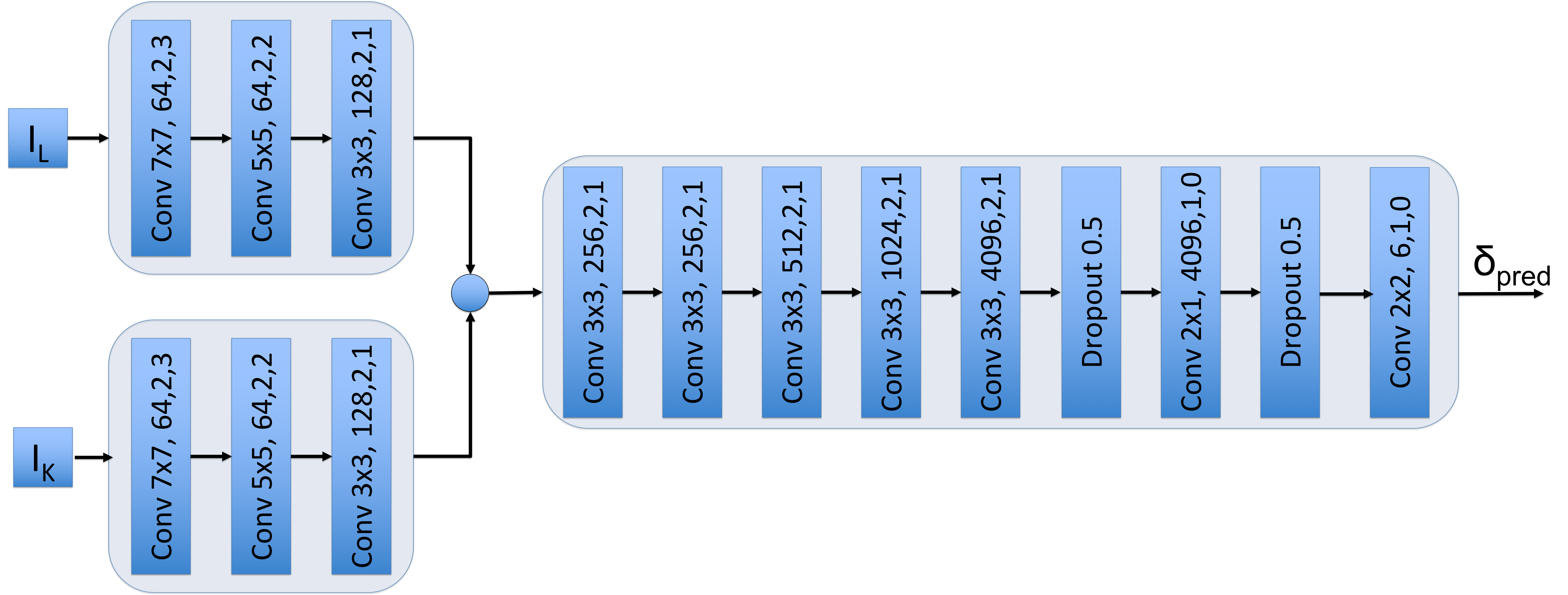}
	\caption[]{Siamese CNN architecture for relative pose regression. The siamese network branches takes in pairs of a keyframe $I_K$ and a live frame $I_L$ and outputs a 6-DoF relative pose $\delta_{pred}$ between those two frames. The Siamese network is a Fully Convolutional Network (FCN) built solely with convolutional layers followed by PReLU non-linearity. Moreover, there are one dropout layer before each of the last two layers.} 
	\label{fig-siamese} 
\end{figure}

Figure~\ref{fig-siamese} shows the network adopted in this work. All convolutional layers, with the exception of the last three, are followed by a non-linearity, PReLUs \cite{He}. The major differences to the network proposed by Handa et al. \cite{Handa2016} are the dropout layers \cite{Srivastava2014} added before the last two layers and a slight larger receptive field in earlier layers. Dropout was chosen for regularization, since the original network \cite{Handa2016} was trained on synthetic data and does not generalize. The receptive field of early layers were made larger, because it is desired to filter out high frequency components of real-world data.

\subsubsection{Loss Function}

Perhaps the most demanding aspect of learning camera poses is defining a loss function that is capable of learning both position and orientation. Kendall et al. \cite{Kendall2015a} noted that a model which is trained to regress both the position and orientation of the camera is better than separate models supervised on each task individually. They first proposed a method to combine position and orientation into a single loss function with a linear weighted sum. However, since position and orientation are represented in different units, a scaling factor was needed to balance the losses. Kendall and Cipolla \cite{Kendall2017b} recommend the reprojection error as a more robust loss function. The reprojection error is given by the difference between the 3D world points reprojected onto the 2D image plane using the ground truth and predicted camera pose. Instead, we chose the 3D projection error of the scene geometry based on the following two reasons. First, it is a representation that combines rotation and translation naturally in a single scalar loss, similar as to the reprojection error. Second, the 3D projection error is expressed in the same metric space as the camera pose and, thus, provides more interpretable results than the reprojection error, which compares a loss function in pixels and a error in meters.

Basically, a 3D projection loss converts rotation and translation measurements into 3D point coordinates, as defined in Equation~\ref{eq:loss},

\begin{equation}\label{eq:loss}
\mathcal{L} = \frac{1}{w \cdot h}\sum_{u=1}^{w}\sum_{v=1}^{h} \norm{ T_{pred} \cdot p(x)^T - T_{gt} \cdot p(x)^T }_2,
\end{equation}

where $x$ is a homogenized 2D pixel coordinate in the live image, $p(x)$ is the $4\times1$ corresponding homogenized 3D point, which is obtained by projecting the ray from the given pixel location $(u, v)$ into the 3D world using inverse camera projection and live depth information, $d(u, v)$, at that pixel location. The norm $\norm{\cdot}_2$ is the Euclidean norm applied to $w \cdot h$ points, where $w$ and $h$ are the depth map width and height, respectively. Moreover, it is worth mention that the intrinsic camera parameters are not required to compute the 3D geometry in the loss function described by Equation~\ref{eq:loss}. The reason is that the same projection $p(u, v)$ is applied to both prediction and ground truth measurements.

Therefore, this loss function naturally balances translation and rotation quantities, depending on the scene and camera geometry. The key advantage of this loss is that it allows the model to vary the weighting between position and orientation, depending on the specific geometry in the training image. For example, training images with geometry far away from the camera would balance rotational and translational error differently to images with geometry very close to the camera. If the scene is very far away, then rotation is more significant than translation and vice versa \cite{Kendall2017b}.

\begin{figure}[ht]
	\centering
	\includegraphics[height = 6cm, width = 0.5\textwidth,keepaspectratio]{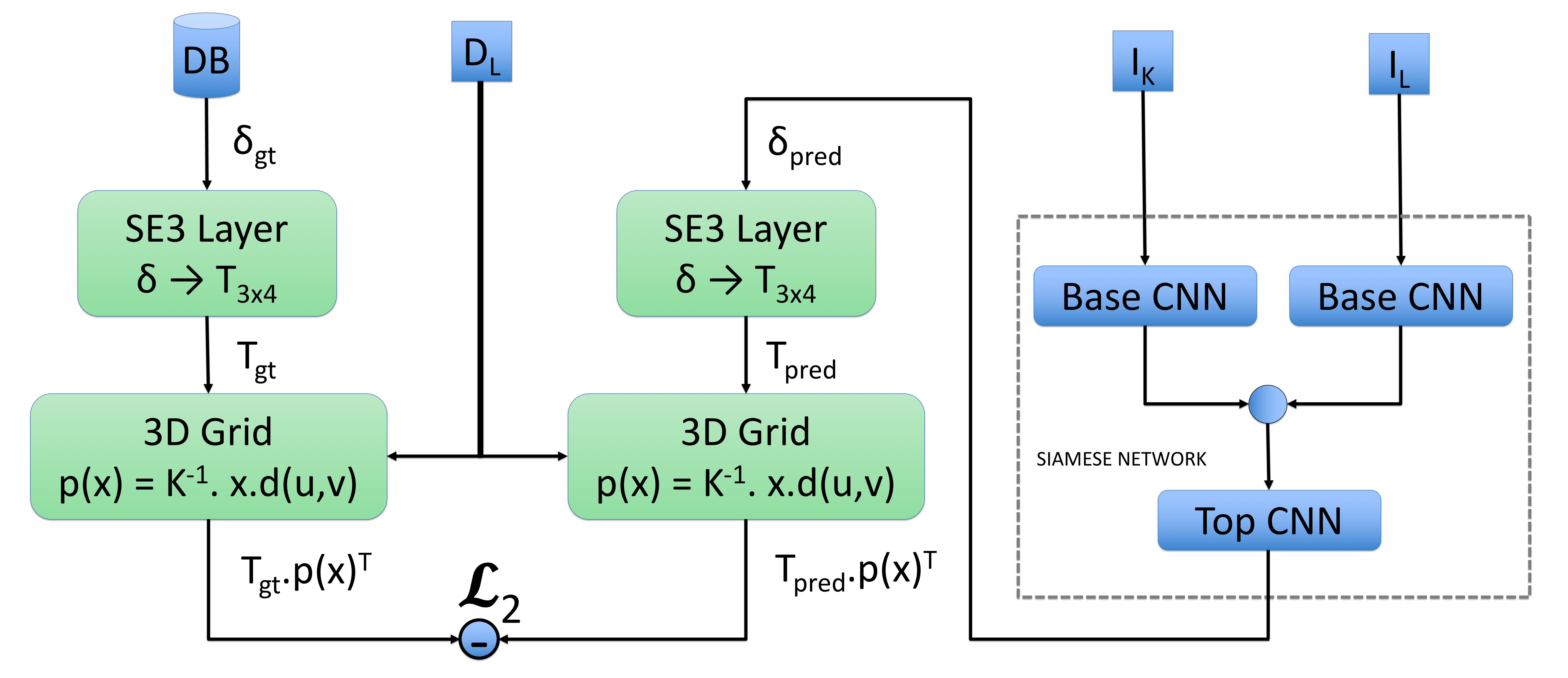}
	\caption[]{Convolution and Geometry layers jointly applied for learning relative camera poses. On top left, it is shown the base and top branches of the siamese network. Its predicted vector $\delta_{pred}$ is passed to the SE3 Layer that outputs a transformation matrix $T_{pred}$. Then, the 3D Grid Layer receives as input a live depth map $D_L$, the camera intrinsic matrix $K$ and $T_{pred}$. Subsequently, it projects the ray at every pixel location $(u,v)$ into the 3D world and multiplies the result by the corresponding live depth $d(u,v)$. Finally, the resulting homogenized 3D point is transformed by the predicted matrix $T_{pred}$. The ground truth relative pose is also passed through the SE3 Layer and the resulting transformation matrix $T_{gt}$ is applied to the output of the 3D Grid Layer produced before, in order to get a view of the 3D point cloud according to the ground truth.} 
	\label{fig-network} 
\end{figure}

The projecting geometry  \cite{Hartley2004} applied to neural network models consists of a differentiable operation that involves matrix multiplication. Handa et al. \cite{Handa2016} provide a 3D Spatial Transformer module that explicitly defines these transformations as layers with no learning parameters. Instead, it allows computing backpropagation from the loss function to the input layers. Figure~\ref{fig-network} illustrates how the geometry layers fit the siamese network. There is a relative camera pose, either given by the siamese network or by the ground truth. There is also geometry layers to compute 3D world point from both relative camera pose and ground truth. On the top left, it is shown the base and top branches of the siamese network that receives as input a pair of a live frame $I_L$ and a keyframe $I_K$ and outputs a relative camera pose vector $\delta_{pred}$. This predicted vector is then passed to the SE3 Layer, which outputs a transformation matrix $T_{pred}$. Following, the 3D Grid Layer receives as input a live depth map $D_L$, the camera intrinsic matrix $K$ and $T_{pred}$. Subsequently, it projects the ray at every pixel location $(u,v)$ into the 3D world (by multiplying the inverse camera matrix $K^{-1}$ by the homogenized 2D pixel coordinate $(u,v)$) and multiplies the result by the corresponding live depth $d(u,v)$. Finally, the resulting homogenized 3D point is transformed by the relative camera transformation encoded in the predicted matrix $T_{pred}$. The ground truth relative pose is also passed through the SE3 Layer and the resulting transformation matrix $T_{gt}$ is applied to the output of the 3D Grid Layer produced before, in order to get a view of the 3D point cloud according to the ground truth.

\subsection{Global Localization System}\label{sec:global_localize}

\begin{figure}[ht]
	\centering
	\includegraphics[height = 3cm, width = 0.5\textwidth,keepaspectratio]{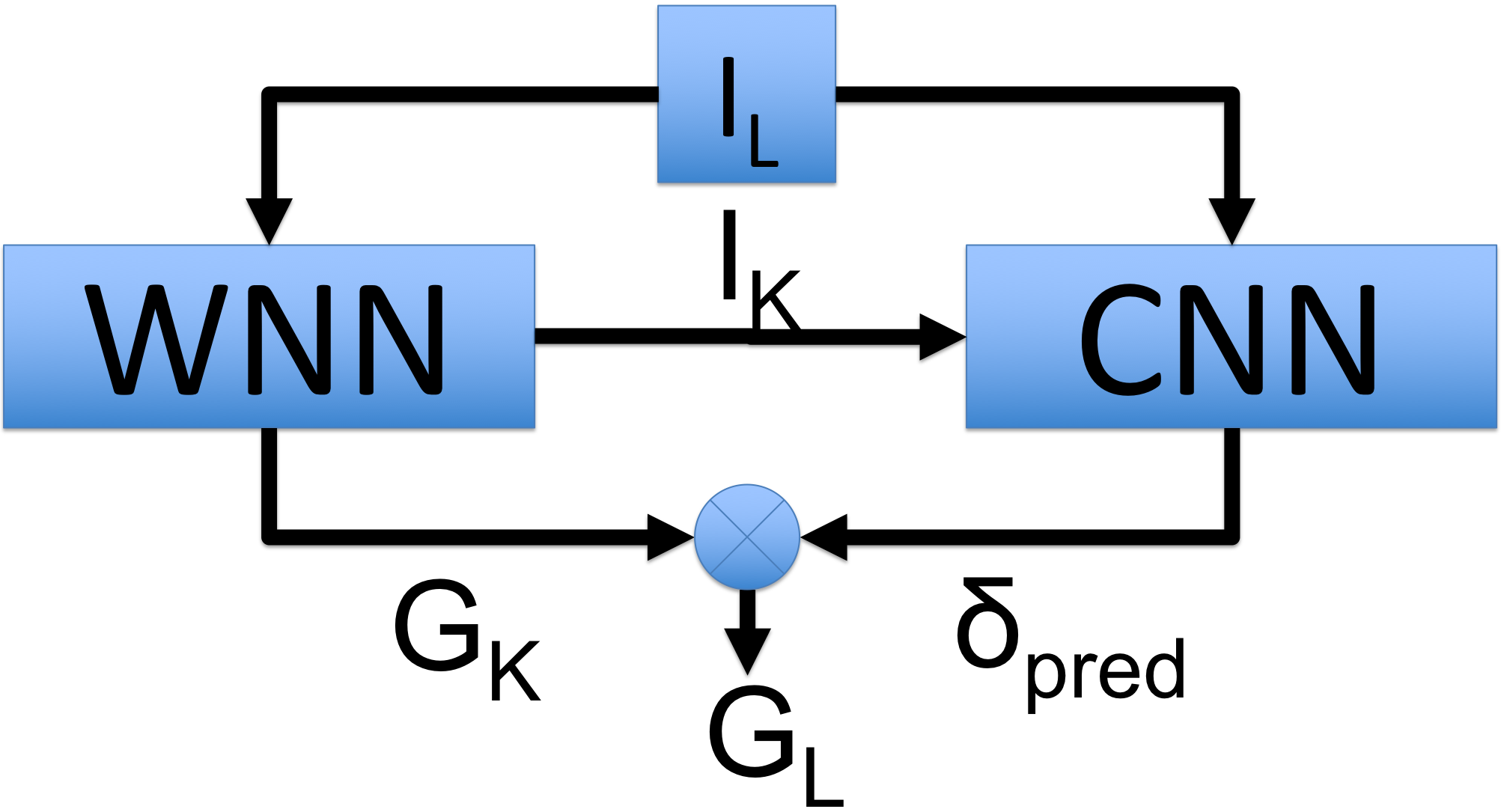}
	\caption[]{The combined WNN-CNN system. The live image $I_L$ is the only input to the whole  WNN-CNN system, which outputs the corresponding live global pose $G_L$. The WNN subsystem outputs the keyframe image $I_K$, which, together with the live image $I_L$, are input to the CNN subsystem, which outputs the relative camera pose $\delta_{pred}$. The last is applied to the key global pose $G_K$ to give the live global pose $G_L$.} 
	\label{fig-system} 
\end{figure}

Lastly, it is presented the integration of the system described in Section~\ref{sec:place_recog} for solving the place recognition problem with the system described in Section~\ref{sec:visual_localize} for solving the visual localization problem. Together, both systems solve the global localization problem, which consists of inferring the current live camera pose $G_L$ given just a single live camera image $I_L$. Note that the only input to the whole system is the live image, as depicted by the smallest square in Figure~\ref{fig-system}.

Figure~\ref{fig-system} shows the workflow between the place recognition system (WNN approach) and visual localization system (CNN approach), which work together to provide the live global pose $G_L$ given the live camera image $I_L$. Live image $I_L$ is sent to both WNN and CNN subsystems, while the WNN recollected image is passed only to the CNN subsystem as the keyframe image $I_K$. Given the image pair, the CNN subsystem outputs the relative camera pose $\delta_{pred}$, which is applied to the key global pose $G_K$, in order to give the live global pose $G_L$.

\section{Experimental Methodology} \label{sec:methodology}		

This section presents the experimental setup used to evaluate the proposed system. It starts describing the autonomous vehicle platform used to acquire the datasets, follows presenting the datasets themselves, and finishes describing the methodology used in the experiments.

\subsection{Autonomous Vehicle Platform}

The data used to evaluate the performance of the proposed system was collected using the Intelligent and Autonomous Robotic Automobile – IARA (Figure~\ref{fig-IARA}). IARA is an experimental robotic platform with several high-end sensors based on a Ford Escape Hybrid that is currently being developed at the Laborat\'orio de Computa\c c\~ao de Alto Desempenho – LCAD (acronym in Portuguese for High-Performance Computing Laboratory) of the Universidade Federal do Esp\'irito Santo (UFES) in Brazil. For details about IARA specifications please refer to \cite{LyrioJunior2014, LyrioJunior2015}.

The datasets used in this work were built using IARA's frontal Bumblebee XB3 stereo camera to capture VGA-sized images at 16fps, and IARA's localization module \cite{Veronese2016} to capture associated poses (6 Degrees of Freedom – 6-DoF). IARA's localization module is based on a Monte Carlo Localization (MCL) \cite{Thrun2005} with an Occupancy Grid Mapping (OGM) \cite{Elfes1989} built with cell grid resolution of 0.2m, as detailed in \cite{Mutz2016}. Poses computed by IARA's Monte Carlo Localization - Occupancy Grid Mapping (MCL-OGM) system has the precision of about the grid map resolution, as verified in \cite{Veronese2016}. 

\subsection{Datasets}

For the experiments, it was collected several laps data in different dates. For each lap, IARA was driven at speeds up to 60 km/h around UFES campus. An entire lap around the university campus has an extension of about 3.57 km. During laps, both image and pose data of IARA were acquired synchronously, amounting to more than 75 thousand pairs of image and pose. Table~\ref{tab:ufes-dataset} summarizes all laps data in ten different sequences. Sequence 8 accounts for two laps, laps 9 and 10 are partial laps and all the others are full laps. The difference in days between sequence 1 and 10 covers more than two years. Such time difference resulted in a challenging testing scenario since it captured substantial changes in the campus environment. Such changes include differences in traffic conditions, number of pedestrians, and alternative routes taken due to obstructions on the road. Also, there were substantial infrastructure modifications of buildings alongside the roads in between dataset recording. The complete set of sequences selected for the experiments is called UFES-LAPS and can be downloaded from the following link \url{https://github.com/LCAD-UFES/WNN-CNN-GL}.

\begin{table}[]
	\centering
	\caption{Ufes Dataset Sequences}
	\label{tab:ufes-dataset}
	\begin{tabular}{lcrrr}
		\toprule
		\multirow{2}{*}{Lap Sequence} & Lap Date &  \multicolumn{3}{c}{Lap Sampling Spacing}\\
		& (mm-dd-yyyy) & None      & 1m      & 5m   \\
		\midrule
		UFES-LAP-01 & 08-25-2016 03:31 PM &    7,165 & 2,868 &     682 \\
		UFES-LAP-02 & 08-25-2016 03:47 PM &    6,939 & 2,726 &     679 \\
		UFES-LAP-03 & 08-25-2016 04:17 PM &    6,404 & 2,663 &    680 \\
		UFES-LAP-04 & 08-30-2016 05:40 PM &    1,808 & 725 & 	  170\\
		UFES-LAP-05 & 10-21-2016 04:15 PM &    9,405 & 2,855 &    669 \\
		UFES-LAP-06 & 01-19-2017 07:23 PM &    1,869 & 704 & 	  171\\
		UFES-LAP-07 & 11-22-2017 05:20 PM &    7,965 & 2,832 &    665 \\
		UFES-LAP-08 & 12-05-2017 09:35 AM &  17,935 & 6,012 & 1,398 \\
		UFES-LAP-09 & 01-12-2018 04:30 PM &    7,996 & 2,868 &    669 \\
		UFES-LAP-10 & 01-12-2018 04:40 PM &    7,605 & 2,899 &    662 \\
		\bottomrule
		\textbf{TOTAL} & & \textbf{75,091} & \textbf{27,152} & \textbf{6,445} \\
	\end{tabular}
\end{table}

To validate the proposed system for both place recognition and visual localization problems, a set of experiments was run with the UFES-LAPS dataset mentioned above. The UFES-LAPS was further split into training, validation and test datasets. The training dataset is named UFES-LAPS-TRAIN and comprises all sequences from UFES-LAPS but the following three: UFES-LAP-04, UFES-LAP-06, and UFES-LAP-07. The UFES-LAP-07 sequence is renamed to UFES-LAPS-TEST to be used for the test. The remaining two sequences, UFES-LAP-04 and UFES-LAP-06, make up the validation dataset, called UFES-LAPS-VALID. This way, UFES-LAPS-TRAIN dataset is used for training, the UFES-LAPS-VALID is used during CNN training to select the best model. Lastly, the UFES-LAPS-TEST dataset is used to test the accuracy of the whole system. 

The dataset sequences were sampled at different sampling spacing. For training, a 5-meter spacing is considered for sampling the sequences from UFES-LAPS-TRAIN, and, for the test, it is considered a 1-meter spacing for sampling sequences from UFES-LAPS-TEST, respectively. In other words, the experiments use a 5-meter spacing UFES-LAPS-TRAIN dataset for training, then it will be called UFES-LAPS-TRAIN-5M. While a 1-meter spacing UFES-LAPS-TEST dataset is used for test and called UFES-LAPS-TEST-1M. The same procedure applies to validation dataset, resulting in the following datasets, respectively: UFES-LAPS-VALID-5M and UFES-LAPS-VALID-1M.

To validate the proposed system for the place recognition problem, a set of experiments was run using the UFES-LAPS-TRAIN-5M and UFES-LAPS-TEST-1M datasets for, respectively, training and test the weightless network. In order to validate the proposed system for the visual localization problem is trained with the keyframes selected from UFES-LAPS-TRAIN-5M, while the live frames are picked from UFES-LAPS-TRAIN-1M. The same procedure applies to validation and test dataset, resulting in the following datasets, respectively: UFES-LAPS-VALID-5M/1M and UFES-LAPS-TEST-5M/1M. To define the ground-truth label between places, the correspondences between every two lap data were established using the Euclidean distance between pairs of image-pose from each lap of training and test datasets with a third dataset, for pose registration purposes only. So the UFES-LAP-05 sequence was reserved for pose registration only and none of its images were considered for place recognition. Firstly it was sampled at the fixed 1m spacing interval to create UFES-LAPS-REG-1M. Following, the UFES-LAPS-TRAIN-1M dataset was matched with the registration dataset UFES-LAPS-REG-1M using the Euclidean distance as proximity measure. Finally, the same procedures are applied to the UFES-LAPS-TEST-1M dataset. The final sizes of registered training and test datasets for place recognition are 4,415 and 2,784, respectively.

In order to define the ground-truth relative vector between camera poses, the relative distances between every two sequence data were established using the Euclidean distance between pairs of a key- and live- frames along with their corresponding poses. The UFES-LAPS-TRAIN-1M dataset was matched with the UFES-LAPS-TRAIN-5M using the Euclidean distance as a proximity measure to select the closest keyframe from the 5m spacing dataset. The same procedure applies to the UFES-LAPS-TEST-1M/5M and UFES-LAPS-VALID-1M/5M datasets. The crossing data combinations for each dataset is as follows. For training data, it is crossed the data of every live frame in UFES-LAPS-TRAIN-1M with the keyframes in UFES-LAPS-TRAIN-5M. For the validation data, live frames come from sequence data in UFES-LAPS-VALID-1M dataset while the keyframes can be in any sequence data from UFES-LAPS-VALID-5M or UFES-LAPS-TRAIN-5M dataset. The same procedure applies to the test dataset. Select the live frames from sequence data in UFES-LAPS-TEST-1M and the keyframes from sequence data in UFES-LAPS-TEST-5M or UFES-LAPS-TRAIN-5M dataset. The final sizes after crossing the sequence data of training, validation and test datasets are, respectively: 98,404, 3,471 and 14,249. 

\subsection{Network Training}

In this subsection, it is described the training procedure and parameter selection for WNN and CNN. For the WNN, the parameters were chosen accordingly to tuning parameter selection done in \cite{LyrioJunior2014} as follows: one neural layer with size $N = 96 \times 54$, where each neuron reads a binary feature vector with size $S = 128$ from the input layer.

The WNN is trained on the UFES-LAPS-TRAIN-5M dataset using images from the left camera of Bumblebee XB3 cropped to $640\times364$. The same crop window applies for UFES-LAPS-TEST-1M. For more details about the weightless network parameters and training procedure please refer to \cite{LyrioJunior2014} . The CNN is trained on the UFES-CNN-LAPS-TRAIN-5M/1M dataset using images from left camera of Bumblebee XB3 and the depth image computed with SPS stereo \cite{Yamaguchi}, being both image and depth cropped to $320\times240$. No data augmentation was used. 

The CNN was trained with Adam optimizer \cite{Kingma2014} using mini-batches of size 24. Adam hyper-parameters $\beta_1$  and $\beta_2$ were set to 0.9 and 0.999, respectively. The learning rate is initially set to $0.0001$ and decreased by a factor of 2 at each epoch. The network was trained for 7 epochs, with 4,101 iterations per epoch. To prevent the network from overfitting, it is employed Dropout layers \cite{Srivastava2014} and Early Stopping \cite{Goodfellow2016}. There are two Dropout layers in the convolutional network architecture presented in Section~\ref{sec:visual_localize}. Both have probability $p=50\%$ of units being randomly dropped at each training iteration. Following early stopping criteria, the training was interrupted and the best model, which achieves smaller positioning error on validation data, was saved.

\begin{figure}[ht]
	\includegraphics[height = 6cm, width = 0.5\textwidth,keepaspectratio]{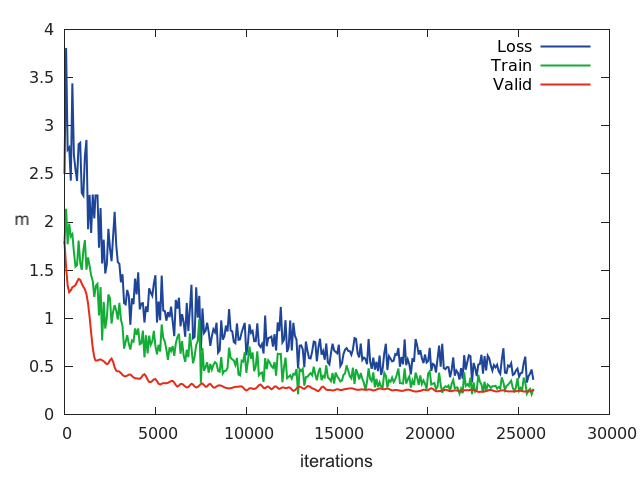}
	\caption[]{CNN training. The vertical axis represents the error in meters and the horizontal axis represents the training iterations. The curves in indigo, green and red represents the error measured in meters of the loss function, training and validation data. The loss function error is defined as in Equation~\ref{eq:loss} and represents the mean Euclidean distance between the ground truth and the predicted 3D point projections, while the training and validation metrics measures the mean Euclidean distance between the 3D camera position given by the network and the ground truth.} 
	\label{fig-training} 
\end{figure}

The curves of the graph in Figure~\ref{fig-training} show the CNN training evolution using UFES-LAPS-TRAIN-5M/1M dataset for training and UFES-LAPS-VALID-5M/1M dataset for validation. The vertical axis represents the error in meters and the horizontal axis represents the number of iterations. The curve in indigo presents the loss function error as in Equation~\ref{eq:loss}, while the curve in green presents the positioning error measured with the Euclidean distance on the training data. The curve in red is also measured with the Euclidean distance but represents the positioning error of validation data.

As the graph of the Figure~\ref{fig-training} shows, the loss function curve stays consistently above all others, while the validation error curve crosses the training error curve after 25,000 iterations. What indicates that the network is overfitting. Following early stopping criteria, the training was interrupted and the best model, which achieves smaller positioning error on validation data, was saved.

\section{Results and Discussions} \label{sec:results}

This section shows and discusses the outcomes of the experiments. It starts describing the performance of the WNN subsystem performance in terms of classification accuracy and follows presenting the CNN subsystem performance in terms of positioning error. A demo video, that shows the WNN-CNN system performance on the UFES-LAPS-TEST-1M test dataset, is available at \url{https://github.com/LCAD-UFES/WNN-CNN-GL}.

\subsection{Classification Accuracy}	

This subsection compares the performance of the WNN subsystem by means of the relationship between the number of frames learned by the system and its classification accuracy. The system classification accuracy is measured in terms of how close the estimated image-pose pair is to the ground-truth image-pose pair. 

\begin{figure}[htbp]
	\includegraphics[width = 0.5\textwidth,keepaspectratio]{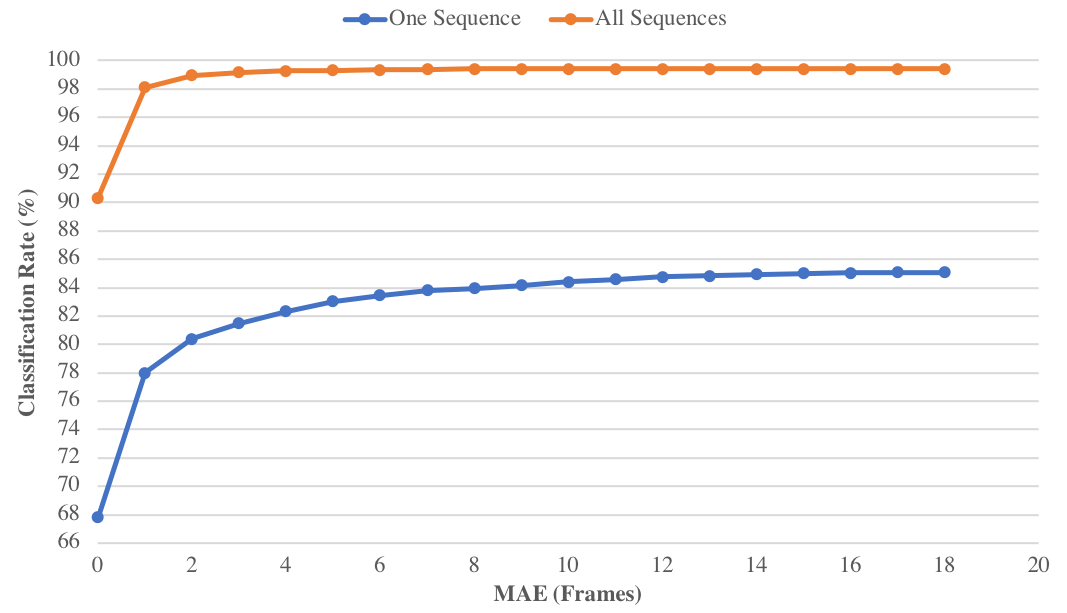}
	\caption{Classification accuracy of VG-RAM WNN for different Maximum Allowed Error (MAE) in frames when training with one sequence (UFES-LAP-01-5M) or with all sequences (UFES-LAPS-TRAIN-5M) and test with the UFES-LAPS-TEST-1M dataset for both.} 
	\label{fig:accuracy} 
\end{figure}

Figure~\ref{fig:accuracy} shows the classification accuracy results obtained on UFES-LAPS-TEST-1M test dataset and using for training either one sequence UFES-LAP-01 at the fixed 5m spacing or all sequences (UFES-LAPS-TRAIN-5M). The vertical axis represents the percentage of estimated image-pose pairs that were within an established Maximum Allowed Error (MAE) in frames from the ground-truth image-pose pair. The MAE is equal to the amount of image-pose pairs that one has to go forward or backward in the test dataset to find the corresponding query image. The horizontal axis represents the MAE in frames. Finally, the curves represent the results for different training datasets: one sequence or all sequences.

As the graph of Figure~\ref{fig:accuracy} shows, the WNN subsystem classification accuracy increases with MAE for both datasets but reaches a plateau at about 3 frames for the all-sequences dataset and at about 10 frames for the one-sequence dataset. For the latter dataset, if one does not accept any system error (MAE equals zero), the accuracy is about 68\%. But, if one accepts an error of up to 3 frames (MAE equals 3), the accuracy increases to about 82\%. On the other hand, when using the all-sequences dataset for training, the system accuracy increases more sharply. For example, with MAE equals to 1, the classification rate is about 98\%. Although the system might show better accuracy with increasing MAE, the positioning error of the system increases. This happens because one frame of error for the training datasets represents 5 meters. 

When comparing the graph curves of Figure~\ref{fig:accuracy}, it can be observed that, for all-sequences training dataset, the WNN subsystem achieves up to 90.3\% in terms of classification accuracy with MAE equals to 0.

\subsection{Positioning Error}

In this subsection, it is analyzed the performance of CNN given the ground-truth keyframe (GT+CNN) and when the keyframe is outputted by WNN (WNN+CNN). Both are compared against GPS on the UFES-LAPS-TEST-1M dataset, where both WNN and GPS systems are more accurate. As seen before, the WNN subsystem is more accurate 90.3\% of the time, assuming a Maximum Allowed Error (MAE) equals to zero. The GPS subsystem is more accurate where the signal quality is stable, which occurs 89.65\% of the time. For this experiment, a signal is considered stable when GPS quality indicator is greater than 0 \footnote{$http://www.trimble.com/OEM_ReceiverHelp/V4.44/en/NMEA-0183messages_GGA.html$}.

\begin{figure}[ht]
	\includegraphics[height = 6cm, width = 0.5\textwidth,keepaspectratio]{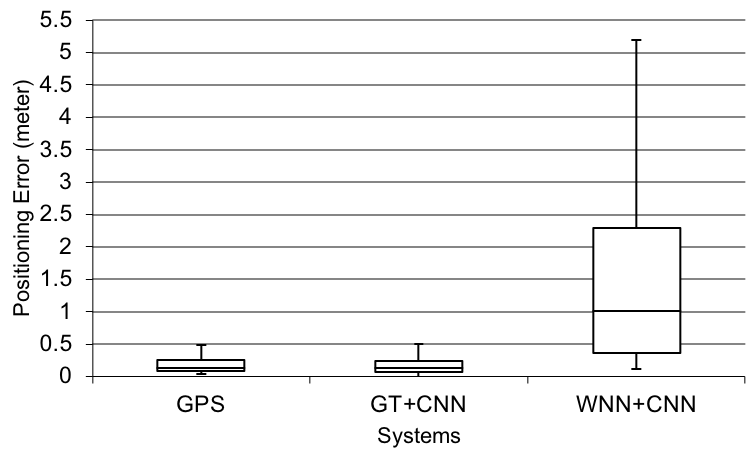}
	\caption[]{Comparison between hybrid WNN-CNN system and GPS positioning error.} 
	\label{fig-boxplot} 
\end{figure}

We measured the positioning error of the proposed system and GPS by means of how close their estimated trajectories are to the trajectory estimated by the OGM-MCL system (our ground truth) on the UFES-LAPS-TEST-1M test dataset. Figure~\ref{fig-boxplot} shows the results as box plots with median, inter-quartile range and whiskers of the error distribution for each system.

As shown in Figure~\ref{fig-boxplot}, positioning errors of GPS and GT+CNN systems are equivalent. For the WNN+CNN system, the positioning error of 50\% of the poses are under 1m and of 75\% of the poses are under 2.3m. Extending the comparison to the Visual SLAM system \cite{LyrioJunior2015} in a similar context, the combined approach has mean positioning error of 1.20m, slightly higher than the 1.12m performed by the Visual SLAM system in the same trajectory. Considering they serve different purposes, combined results of the hybrid WNN-CNN approach looks promising.

\section{Conclusions and Future Work}\label{sec:conclusion}

It was shown that to solve the global localization problem, it is required more than just outputting the position where the robot was during mapping phase. It is desired to approximate the actual robot position and orientation with respect to the past pose. This problem was tackled here by training a Siamese-like CNN that takes as input two images and regresses a 6-DoF relative pose. It was advocated that using a geometry loss to project the 3D points transformed by network's output pose is a better approach than using the ground truth pose as the backpropagation signal. It naturally balances the differences in the scaling of the position and rotation units, for instance. It was also verified that the loss function error is consistently above training and validation errors. The geometry loss function apparatus demonstrated being a robust loss for the task of regressing the relative pose. For the final experiment, the best-trained model was applied in conjunction with the WNN subsystem to solve the global localization problem. It was shown that the combined results of the hybrid WNN-CNN approach were on pair with a Visual SLAM system, although needs improvements compared to RTK-GPS precision. 

Some direction for future work involves extending this work with larger datasets and evaluating the network performance using transfer learning and fine tuning with Ufes dataset. As more data are provided \cite{Contreras2017}, it is expected an increase in accuracy and regularization for \textit{Deep Learning} models. For instance, the PoseNet's \cite{Kendall2015a} localization accuracy was improved by increasing the number of training trajectories, while maintaining a constant-size CNN. Conversely, for the WNN subsystem, larger datasets can degrade runtime performance as the runtime during test scales with the number of training samples.  

\textit{Deep Learning} models have demonstrated superhuman performance in some tasks but at the expense of large amounts of correctly labeled data for training models using standard supervised techniques,  which is costly in robotics. To overcome this issue, an alternative is to train \textit{Deep Learning} models using weakly supervised techniques \cite{Zhou2017d}  with noisy labeled data, or even unlabeled data. This could open doors to many new applications in robotics, such as Visual SLAM using end-to-end \textit{Deep Learning} techniques. For the relative pose estimation problem studied here, another alternative to overcome noisy labeled data is to incorporate its uncertainty in the loss function as an extra parameter \cite{Kendall2017b}.

%Another direction for future research is to experiment recent new models, such as Capsule Networks \cite{Sabour2017} and Gated Feature Learning models \cite{Memisevic2013}. The latter uses multiplicative interactions for learning content-independent features, in order to encode only the transformation between image pairs. These models learn content-independent features that encode the transformation between image pairs. So, it would be interesting to verify whether they can learn to relate pairs of keyframes and live frames, like those used to train the CNN in Section~\ref{sec:visual_localize}, by means of a transformation between them, instead of their content.

\section*{Acknowledgment}

The authors would like to thank NVIDIA Corporation for the donation of some GPUs used in this work and Conselho Nacional de Desenvolvimento Cient\' ifico e Tecnol\' ogico – CNPq, Brazil (grants 311120/2016-4 and 311504/2017-5) for their financial support to this research work.

\bibliographystyle{IEEEtran}
\bibliography{library}

\end{document}